\documentclass[journal]{IEEEtran}

\usepackage{cite}
\usepackage{graphicx}
\usepackage[cmex10]{amsmath}
\usepackage{array}
\usepackage{float}
\usepackage{stfloats}
\usepackage[hidelinks]{hyperref}
\usepackage[numbers,sort&compress]{natbib}
\setcitestyle{numbers,square,sort&compress}

\usepackage{booktabs}
\usepackage{tabularx}
\usepackage{ragged2e}
\usepackage{amssymb}
\usepackage[table]{xcolor}
\usepackage{placeins}
\usepackage{balance}
\usepackage{url}

\definecolor{DeltaRow}{gray}{0.88}
\newcommand{\deltaval}[1]{#1}

\begin{document}

\title{PointRL: Learning Point-Level Vision-Language Grounding from Verifiable Annotation Evidence}

\author{Jingyang~Su,~Pu~Cao,~Xiuze~Jin,~Longyue~Zhang,~Qing~Song,~and~Lu~Yang\thanks{School of Intelligent Engineering and Automation, Beijing University of Posts and Telecommunications, Beijing, China\\
No. 1 Nanfeng Road, Shahe Higher Education Park, Shahe Area, Changping District, Beijing, 102206, China\\
Emails: 2025111367@bupt.cn, caopu@bupt.edu.cn, jxzzz647aa@bupt.edu.cn, zhanglongyue@bupt.edu.cn, priv@bupt.edu.cn, soeaver@bupt.edu.cn}\thanks{Corresponding author: soeaver@bupt.edu.cn}}

\IEEEoverridecommandlockouts
\maketitle

\begin{abstract}
Vision-language models (VLMs) increasingly rely on point coordinates as a compact and executable interface for visual grounding in GUI interaction, robotic manipulation, and interactive visual systems. However, learning reliable pointing behavior remains difficult because the supervision space is inherently non-unique: many coordinates may be valid within the same target region, while multi-instance instructions require target coverage, count consistency, and duplicate suppression. This work presents PointRL, a verifiable reinforcement learning framework that learns point-level grounding from existing heterogeneous annotation evidence. PointRL converts bounding boxes, masks, and instance labels into pointing instructions, while retaining their target supports, instance membership, and set constraints as hidden verifier evidence, i.e., annotations kept outside the prompt and used by a deterministic checker to score predictions. The proposed reward evaluates parseability, point validity, instance coverage, cardinality consistency, and redundant or missing predictions. On PointArena, PointRL improves the overall accuracy of Qwen3.5-4B from 56.11\% to 65.58\%. Further evaluations on RoboSpatial, BLINK, and Ref-Adv show same-backbone gains on the evaluated external benchmarks, suggesting that verifiable point-level feedback may benefit spatial grounding in these settings.
\end{abstract}


\renewcommand{\IEEEPARstartWORDFONTSTYLE}{\rmfamily}
\renewcommand{\IEEEPARstartWORDCAPSTYLE}{\relax}

\section{Introduction}
\label{introduction}
\IEEEPARstart{V}{ision-language grounding is a fundamental capability of vision-language models (VLMs)}, requiring a model to connect natural-language instructions with objects, regions, and locations in visual scenes. This capability supports a broad range of interactive and embodied applications, including GUI agents, robotic manipulation, assistive perception, and human-in-the-loop visual systems~\citep{kamath2021mdetr,liu2024grounding,hong2024cogagent}. Although grounding is commonly represented by bounding boxes or segmentation masks, point coordinates provide a more compact and directly executable interface. A point can indicate a click position, a manipulation target, or an interactive segmentation prompt, and VLMs can generate such coordinates as text using standard decoding. The prompt-response framing follows instruction-tuning style training~\citep{liu2023visual}.

\begin{figure}[H]
    \centering
    \includegraphics[width=\linewidth]{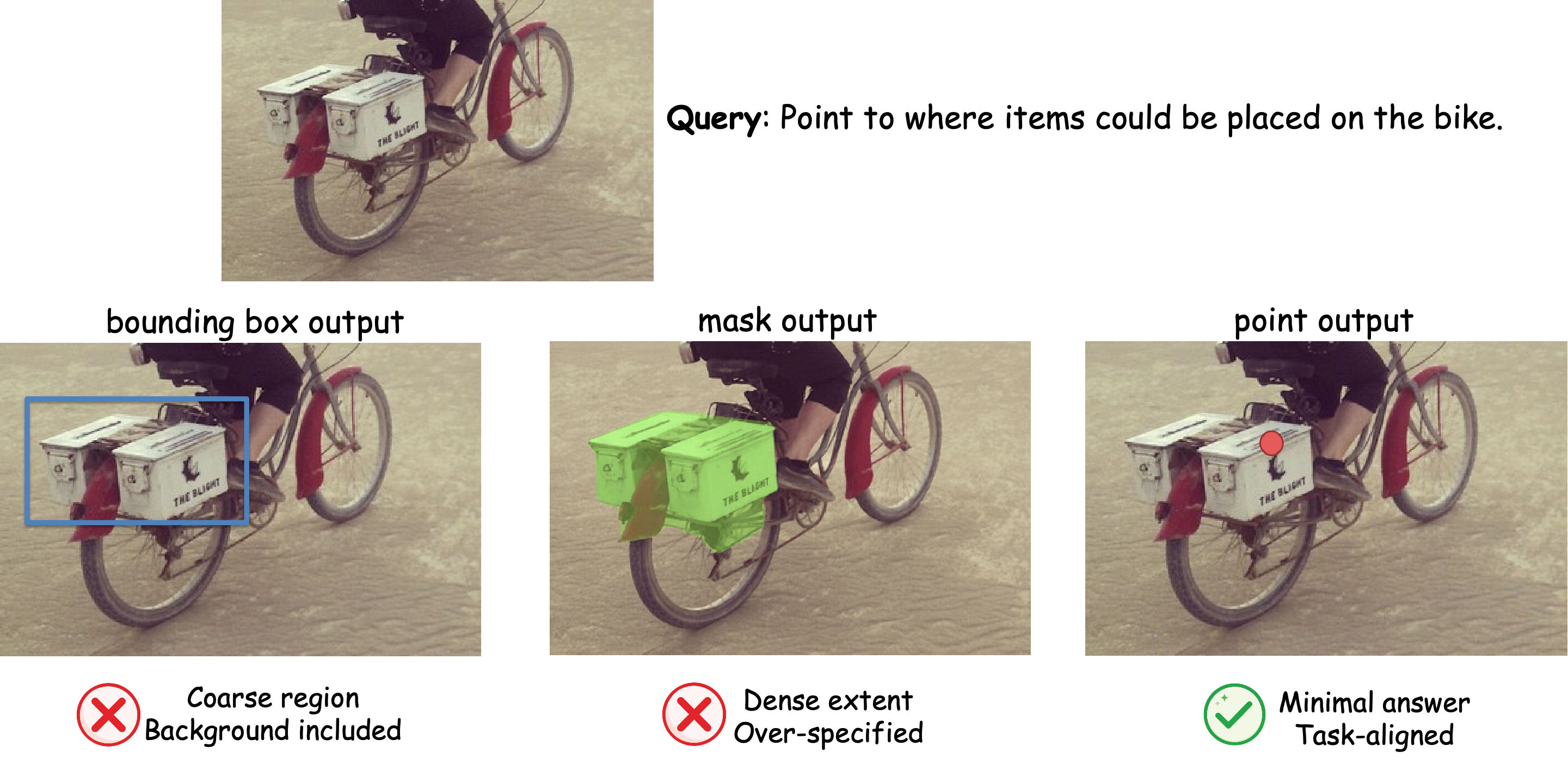}
    \caption{Task illustration for verifiable point-level grounding. A valid response should place points on the intended target supports, while nearby distractors, incomplete coverage, or shortcut outputs are rejected by the verifier.}
    \label{fig:task_intro}
\end{figure}

However, learning reliable pointing behavior goes beyond standard single-label prediction. For a target region, many coordinates may be correct as long as they fall inside the valid spatial support. Reducing such supervision to a single fixed point label unnecessarily collapses a set of valid answers and may penalize equally acceptable predictions. Recent robust visual grounding studies have also emphasized ambiguity-aware supervision~\citep{lu2024lgr,wang2024masked,qiu2024mcce}. The difficulty becomes more pronounced in instruction-conditioned settings. Affordance-oriented queries require the model to identify functionally relevant regions, spatial-relation queries require reasoning over relative positions, and multi-instance queries require a set of distinct points that covers all target instances with the correct cardinality. Therefore, the central challenge is to learn point predictions that satisfy both geometric validity and instruction-specific set constraints.

Existing grounding annotations provide useful evidence for connecting language to visual entities. Bounding boxes and segmentation masks specify target supports on which a target may lie, while instance labels define sets of objects that may satisfy an instruction. PointRL treats these raw annotations as annotation evidence and normalizes them into hidden verifier evidence \(V=(T,C)\), where \(T\) stores the targets and \(C\) stores instruction-conditioned constraints~\citep{xu2024cmirnet,liu2025refersam}. This representation is kept outside the prompt and used as an executable interface for reward computation.

This work introduces \textbf{PointRL} for verifiable reinforcement learning of visual pointing from heterogeneous grounding annotations. PointRL converts boxes, masks, and instance labels into natural-language pointing instructions, while preserving the original annotations as hidden verifier evidence. During training, the VLM generates a textual response that is parsed into one or more coordinates, and a deterministic verifier computes a structured reward from the predicted point collection and \(V\). The reward evaluates output parseability, point validity, localization quality, instance coverage, cardinality consistency, and redundant or missing predictions. In this way, PointRL preserves the non-unique answer space of region-based supervision while handling point annotations through distance-based scoring, and it transforms heterogeneous annotations into executable verifier evidence for point-level learning.

The evaluation asks whether verifier-based point feedback improves same-backbone grounding on PointArena and several external spatial grounding benchmarks. On PointArena, PointRL improves the overall accuracy of Qwen3.5-4B from 56.11\% to 65.58\%. Additional evaluations on RoboSpatial, BLINK, and Ref-Adv show same-backbone gains on the evaluated external benchmarks, suggesting that verifiable point-level feedback may benefit spatial grounding in these settings.

The contributions are threefold. First, PointRL builds prompts and hidden verifier evidence from annotations. This evidence stores target supports, target sets, and instruction-conditioned constraints. Second, it defines a hierarchical reward mechanism that handles non-unique valid points and multi-instance constraints, including localization, coverage, cardinality consistency, and duplicate suppression. Third, same-backbone evaluations on in-domain and external benchmarks provide evidence that verifiable feedback derived from heterogeneous grounding annotations improves point-level grounding in the evaluated settings.

\section{Related Work}
\label{sec:related-work}

\subsection{Visual Grounding and Spatial Evidence}
Visual grounding aims to localize visual entities described by natural language. Early two-stage methods typically generate candidate regions and rank them according to their semantic correspondence with the query, while end-to-end approaches such as MDETR~\citep{kamath2021mdetr} and Grounding DINO~\citep{liu2024grounding} directly predict language-conditioned bounding boxes. Referring-expression grounding has also been studied with modular attention models such as MattNet and neural module-tree approaches~\citep{yu2018mattnet,liu2019learning}. With the development of multimodal large language models, grounding has been extended from box prediction to richer spatial output formats~\citep{lu2024lgr,wang2024masked,qiu2024mcce}. For example, LISA~\citep{lai2024lisa} and GLaMM~\citep{rasheed2024glamm} study mask-based referring segmentation, while LGR-NET~\citep{lu2024lgr}, CMIRNet~\citep{xu2024cmirnet}, RefSAM~\citep{liu2025refersam}, and SSP-SAM~\citep{tang2025ssp} focus on referring image segmentation under richer supervision cues. LocateAnything~\citep{wang2026locateanything} and PaDT~\citep{su2025patch} explore unified spatial outputs through specialized decoding mechanisms or large-scale grounding data engines. AD-DINO~\citep{guo2025ad} further links attention-dynamic reasoning with embodied reference understanding.

Recent studies also emphasize fine-grained spatial understanding in structured scenes, including point-guided localization~\citep{song2020learning}, instance-level human parsing~\citep{yang2019parsing,li2024frequency,guo2026quality}, and omnidirectional person positioning~\citep{yang2025large,yang2023large}. These works show that visual grounding combines localization with spatial structure and relation-aware target selection~\citep{yang2024deep,yang2022survey,zhao2025constructing}. However, existing grounding supervision remains highly heterogeneous, including boxes, masks, instance labels, and relation metadata, which has motivated generalized multi-task grounding formulations~\citep{dai2025gc}. Such heterogeneity makes it difficult to define a unified learning signal across datasets and tasks. PointRL addresses this issue by treating points as a shared prediction interface and by using heterogeneous annotations as hidden verifier evidence that provides spatial and instance-level knowledge for reward computation.

\subsection{Language-Guided Pointing}
Point prediction has recently emerged as a compact interface for language-guided grounding. In this format, a model outputs one or more image coordinates that directly indicate target locations. This format is particularly suitable for interactive systems, because a point can correspond to a click, a manipulation target, or a prompt for downstream segmentation. Molmo and PixMo~\citep{deitke2024molmo} show that modern multimodal models can acquire pointing behavior from large-scale point supervision, including counting-by-point for multi-instance references. MolmoPoint~\citep{clark2026molmopoint} further improves pointing accuracy by introducing grounding tokens and visual-token selection, representing an architecture-oriented approach to point prediction.

Dedicated benchmarks have also been developed to evaluate pointing ability. PointArena~\citep{cheng2025pointarena} formalizes language-guided pointing through Point-Bench and Point-Battle protocols, covering spatial, affordance, counting, steerability, and reasoning-oriented instructions. Pointing also serves as an action interface in robotic manipulation~\citep{yuan2024robopoint} and GUI grounding~\citep{li2025screenspot}, where the predicted coordinate can be executed by an external system. PointRL follows a verifier-based route for learning pointing behavior from existing boxes, masks, and instance labels. The framework preserves verifiable target regions during reward learning and converts them into point-level feedback.

\subsection{Reinforcement Learning with Verifiable Rewards}
Reinforcement learning with verifiable rewards uses rule-based or automatically computable outcome signals as training feedback for model behavior. This paradigm has been demonstrated in language reasoning by DeepSeek-R1~\citep{guo2025deepseek} and has recently been extended to vision-language models. Reward design can follow potential-based invariance principles to avoid changing the optimal policy~\citep{Ng+HR:1999}. Visual-RFT~\citep{liu2025visual} applies reinforcement fine-tuning to visual perception tasks, VLM-R1~\citep{shen2025vlm} adopts GRPO-style optimization for visual reasoning, and GenSeg-R1~\citep{hegde2026genseg} introduces segmentation-quality feedback for referring segmentation.

Most existing visual verifiable rewards are designed around task-level correctness or region-level quality, such as answer accuracy, box overlap, or mask overlap. Point-level grounding requires rewards that handle multiple valid coordinates for the same target region, as well as coverage, count consistency, and duplicate suppression for multi-instance instructions. PointRL defines rewards over unordered point collections. Its verifier-based reward combines soft localization, point-target matching, and set-level constraints, allowing heterogeneous grounding annotations to supervise point prediction while preserving the non-unique answer space for region-based supports.

\section{PointRL-Data: Constructing Verifiable Pointing Data from Grounding Annotations}
\label{sec:data-construction}

\subsection{Task Formulation}
\label{sec:pointing-task}

A pointing task instance consists of an image \(I\), a natural-language instruction \(q\), and hidden verifier evidence \(V=(T,C)\). The instruction specifies the referent or referents that should be pointed to in the image. During execution, the model receives only \(I\) and \(q\), and returns a collection of points,
\begin{equation}
P=(p_i)_{i=1}^{n}, \quad p_i=(x_i,y_i).
\end{equation}
The required size of \(P\) is determined jointly by the instruction and the image content. A query for a single referent calls for one point, whereas a query over all matching instances calls for one point per visible target instance. Because target instances have no canonical order, verification treats \(P\) as an unordered collection: point order is ignored, while repeated coordinates are kept as separate predictions so missing, extra, or duplicated points can be penalized.

Correctness is evaluated with \(V=(T,C)\) by a deterministic checker. The target set \(T=\{t_j\}_{j=1}^{m}\) contains the entries that should be pointed to. Each entry \(t_j\) identifies an intended object, region, or location, and stores its support \(S_j\), the geometric region or point used for hit checking, together with metadata needed to distinguish the intended instance. The constraint set \(C\) records requirements that cannot be checked by membership in the support region alone, such as binding to a reference object, relation constraints, cardinality, uniqueness, and functional restrictions.

A response is correct only when the returned point collection matches both the target entries and the constraints. For a single-target instruction, one point must fall inside the intended support. For multi-instance or constrained instructions, the collection must cover the required distinct targets, have the target-set cardinality, avoid duplicate matches, and satisfy all constraints in \(C\).

This formulation separates the task input from the evidence used to judge the answer. The next subsection describes how we construct the instructions, target entries, and constraints from heterogeneous annotations.

\subsection{Dataset Construction}
\label{sec:instruction-construction}

\begin{figure*}[!ht]
    \centering
    \includegraphics[width=\linewidth]{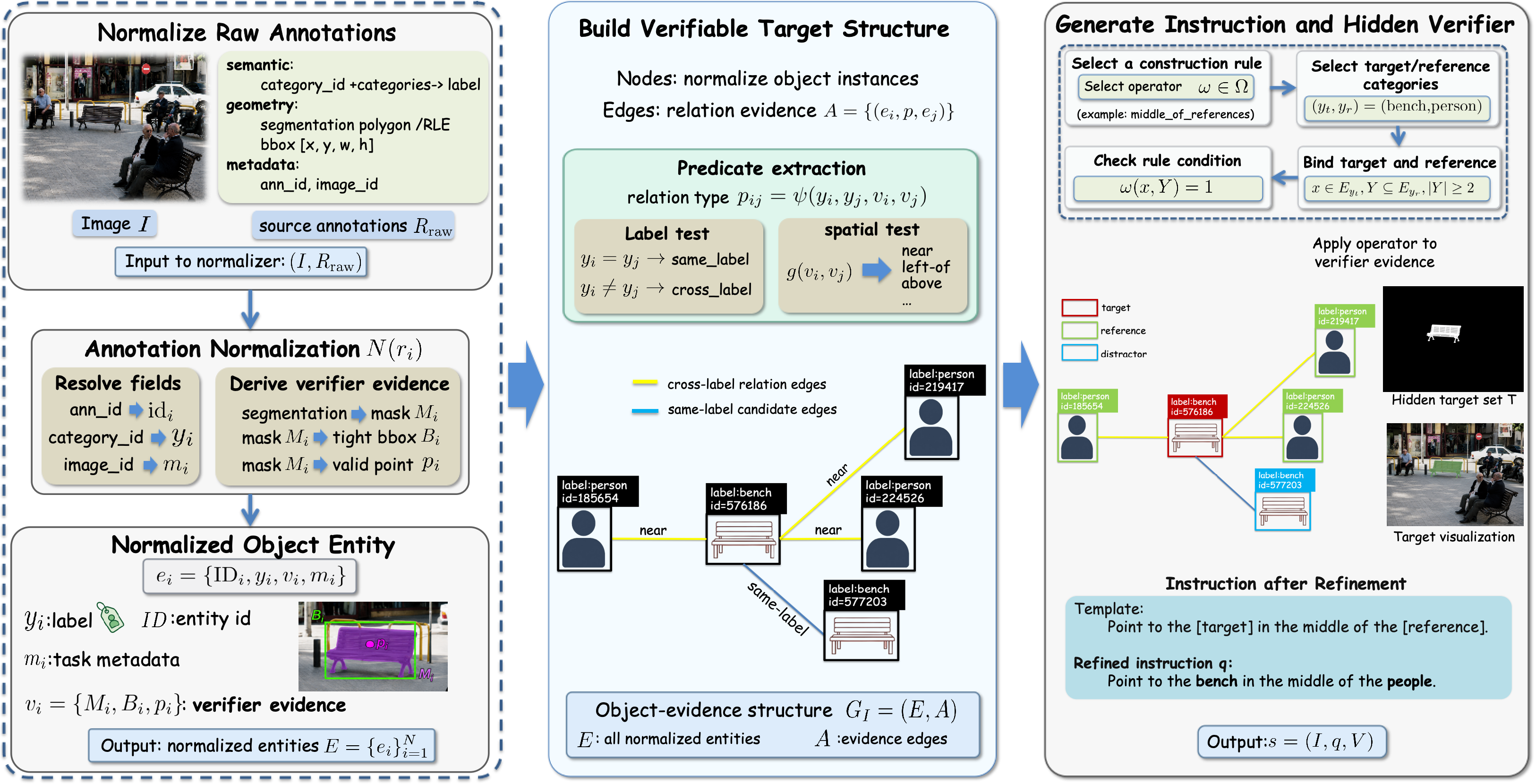}
    \caption{Construction of verifiable pointing samples from raw annotations. The pipeline normalizes annotation sources into object entities, collects supporting evidence, and generates the instruction with hidden verifier evidence
    \(V=(T,C)\). The target set and constraints are then fixed by rule-based checks.}
    \label{fig:dataset_construction}
\end{figure*}

For each image, we convert heterogeneous raw annotations into the verifier interface defined above. The construction pipeline first normalizes source annotations into object entities, then collects the evidence needed to decide which entities can serve as targets or references, and finally applies construction rules to instantiate an instruction \(q\) with hidden verifier evidence \(V=(T,C)\). Here, \(T\) stores the target entries and their supports, while \(C\) stores the conditions used by the deterministic checker. Figure~\ref{fig:dataset_construction} summarizes this pipeline.

Annotation normalization maps source-specific formats into a common entity set \(E=\{e_i\}_{i=1}^{N}\). Each visual referent is represented with an identifier, an optional semantic label, support geometry, and source metadata. The normalized entities are then collected into an object-evidence structure \(G_I=(E,A)\). In this structure, \(E\) lists the object entities, while \(A\) records usable evidence attached to entities or entity pairs, including label agreement, spatial relations, count cues, geometric supports, and functional roles. This entity-centered representation follows prior entity-relation formulations for connecting language and vision~\citep{krishna2017visual}, while keeping the evidence explicit for verification.

Construction rules instantiate \(q\) and \(V\) from \(G_I\), following a rule-based scheme related to programmatic weak supervision for data construction~\citep{ratner2016data,ratner2017snorkel}. Each rule selects a construction pattern, such as category, attribute, relation, counting, region, or functional-role grounding, binds the required target and optional reference entities, checks that the collected evidence supports the instruction, and accepts the sample only when the answer can be reduced to a verified target set. The accepted binding fixes the target supports in \(T\) and records additional requirements in \(C\), such as category membership, attributes, relations, cardinality, uniqueness, region conditions, or functional-role constraints.

Finally, a Qwen-based refiner~\citep{qwen3.5} is used to rewrite the surface form of \(q\) under the accepted rule binding. It is prompted to preserve target categories, counts, reference objects, relations, and functional restrictions, while the verifier fields remain fixed. Before a sample is kept, deterministic checks validate target supports, target unambiguity, constraint consistency, cardinality consistency, and alignment between the rule binding and the hidden verifier evidence. The verifier fields are organized around three requirements:
\begin{itemize}
    \item \textbf{Valid target support:} each target stores a support \(S_j\) for hit checking and soft localization.
    \item \textbf{Required target instances:} multi-target samples store the target set \(T=\{t_j\}_{j=1}^{m}\) and cardinality constraints for unordered matching.
    \item \textbf{Instruction-specific constraints:} relation, functional, reference-anchor, and shortcut-control conditions are stored in \(C\).
\end{itemize}

\subsection{Dataset Statistics}
The accepted samples draw on AGD20K labels~\citep{luo2022learning}, PixMo points~\citep{deitke2024molmo}, COCO masks and instance metadata~\citep{lin2014microsoft}, and rule-bound cue-target pairs. Table~\ref{tab:data-composition} summarizes the resulting rule families. In our experiments, this construction yields 1,647 training samples for reward learning and a 200-sample held-out diagnostic split.

The distribution is intentionally not dominated by a single rule family. Relation and cue-based constructions account for 393 and 398 samples, respectively, or 48.0\% of the accepted set together. These two families stress instruction-conditioned constraints in \(C\), such as reference binding, spatial predicates, and shortcut control. Instance and semantic constructions contribute 345 and 249 samples, covering multi-instance counting, unordered target coverage, and category-level matching. Functional samples add 262 region-oriented examples, which exercise support validity for affordance-style targets. This mixture gives the verifier both geometric evidence, such as masks, boxes, points, and regions, and non-geometric conditions, such as counts, relations, and cue-target constraints.

\begin{table}[H]
\centering
\caption{Composition of accepted PointRL-Data rules.}
\label{tab:data-composition}
\footnotesize
\setlength{\tabcolsep}{3pt}
\renewcommand{\arraystretch}{1.12}
\begin{tabularx}{\linewidth}{@{}l r >{\RaggedRight\arraybackslash}X >{\RaggedRight\arraybackslash}X@{}}
\toprule
Rule & Count & Evidence & Check \\
\midrule
Functional & 262 & AGD regions & Affordance support \\
Instance & 345 & COCO/PixMo instances & Coverage/count \\
Semantic & 249 & Class labels & Category match \\
Relation & 393 & Entity pairs & Spatial relation \\
Cue & 398 & Cue-target pairs & Shortcut guard \\
\midrule
Total & 1,647 & Mixed evidence & Rule-checked targets \\
\bottomrule
\end{tabularx}
\end{table}

\section{PointRL: Reinforcement Learning for Pointing}

PointRL learns point-level grounding by using hidden verifier evidence to score each sampled response. Given an input $z=(I,q)$, the VLM policy samples a textual response $y$. During reward computation, the hidden verifier evidence \(V=(T,C)\) provides the target set \(T=\{t_j\}_{j=1}^{m}\) and instruction-conditioned constraints \(C\). Single-target pointing corresponds to \(m=1\), while multi-target pointing has \(m>1\) and requires the predicted points to cover the target set with the correct cardinality.

\begin{figure*}[!h]
    \centering
    \includegraphics[width=0.95\textwidth]{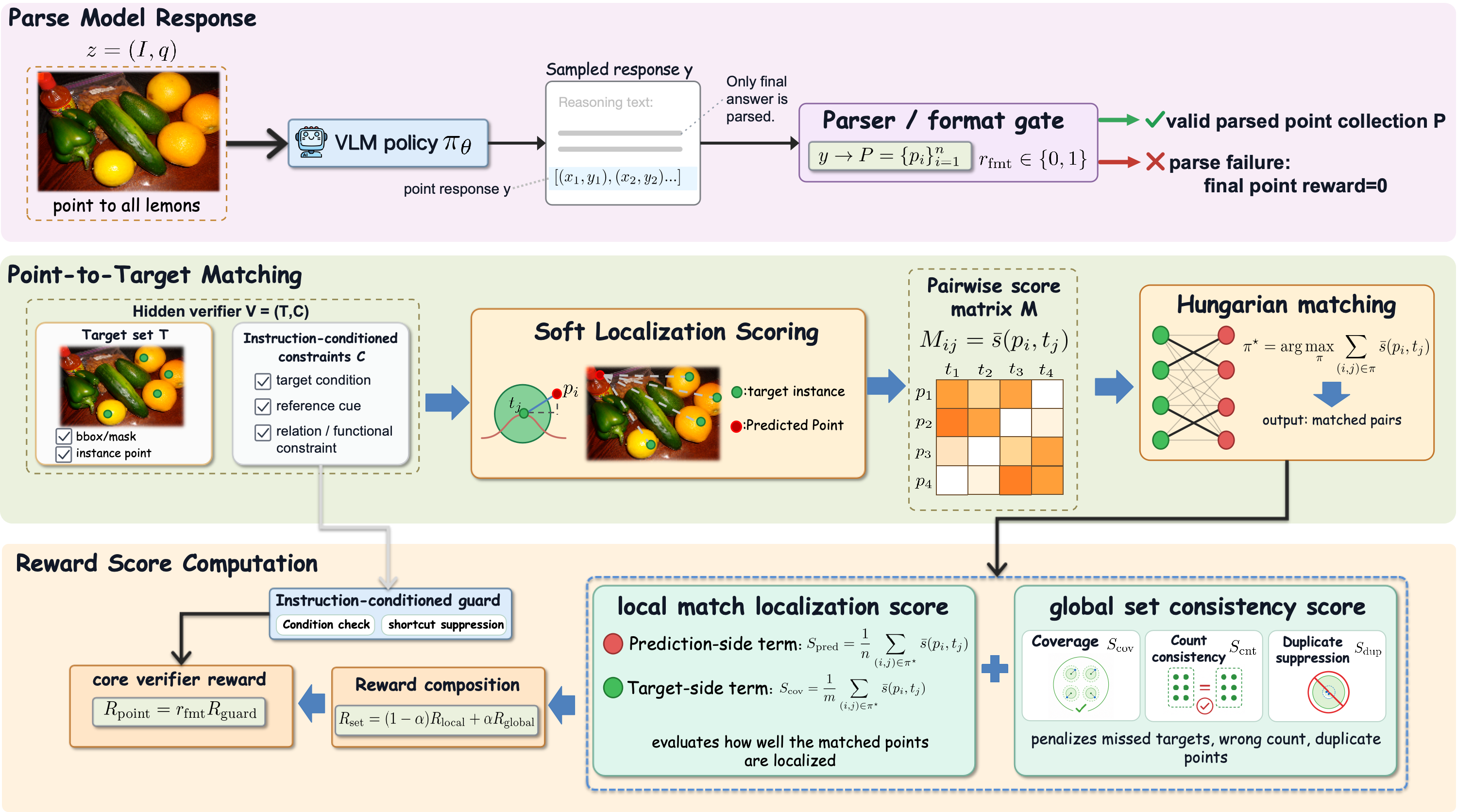}
    \caption{Point reward computation from hidden verifier evidence. The reward module parses the final answer into a point collection, matches predictions to hidden verifier targets with Hungarian matching, and combines matched-pair localization quality with set-level coverage and count checks. Format and instruction-conditioned guards suppress invalid or shortcut responses before returning the core verifier point reward.}
    \label{fig:reward_pipeline}
\end{figure*}

Figure~\ref{fig:reward_pipeline} gives an overview of the reward computation. The reward module first parses \(y\) through a deterministic parser and format gate to obtain the standard point collection \(P\). It then scores \(P\) against \(V\): predicted points are matched to verifier targets, evaluated by local localization quality and global coverage, count, and duplicate consistency, and adjusted by guard terms that suppress instruction-specific shortcuts such as copying a reference cue. These terms are aggregated into the core verifier point reward. Section~\ref{sec:parsing} defines parsing and format validation, Section~\ref{sec:verifier_matching} describes point-to-target matching, and Section~\ref{sec:reward_computation} gives the final reward composition. Implementation-specific serialization is described in Section~\ref{sec:training-details}; all equations in this section operate on the standard point collection \(P\).

\subsection{Point Prediction in VLM}
\label{sec:parsing}
The parser maps the terminal answer block of \(y\) to the standard point collection defined in Eq.~(1), where \(n\) is the number of parsed points. A response is parseable only if the accepted final-answer schema can be normalized to exactly one coordinate list in the form
\[
[(x_1,y_1),\ldots,(x_n,y_n)],
\]
with \(n\ge 1\), two finite numeric coordinates in each tuple, and all coordinates within the image bounds after conversion to the verifier pixel coordinate system. Empty lists, malformed or non-numeric tuples, out-of-bound coordinates, missing point lists, and multiple conflicting final-answer blocks are treated as unparseable.

We set \(r_{\mathrm{fmt}}=1\) for parseable responses and \(r_{\mathrm{fmt}}=0\) otherwise. When \(r_{\mathrm{fmt}}=0\), the final point reward is zero; when \(r_{\mathrm{fmt}}=1\), \(P\) is passed to the hidden verifier evidence \(V=(T,C)\) for matching and reward computation. Duplicate coordinates are retained as separate indexed predictions and handled by the matching and set-level reward terms, including duplicate suppression. The JSON format used in the experiments is an experiment-specific final-answer schema that is normalized back to this point collection interface, as described in Section~\ref{sec:experimental-setup}.

\subsection{Matching Predicted Points to Targets}
\label{sec:verifier_matching}
After parsing a valid point collection \(P=(p_i)_{i=1}^{n}\), the verifier compares these unordered predictions with the target set \(T=\{t_j\}_{j=1}^{m}\). This subsection addresses two steps: scoring each prediction-target pair and assigning predictions to targets without double counting. A binary hit indicator is useful for hard correctness, but it is too sparse as a reward signal: a point just outside the valid target support and a point far from the object would both receive zero credit. PointRL therefore uses the distance from each prediction to each target support as the basis for localization scoring, and then performs one-to-one matching over the resulting score matrix. This subsection focuses on the target supports in \(T\); the remaining instruction-conditioned constraints in \(C\) are applied by the guard terms in Section~\ref{sec:reward_computation}.

Each verifier target \(t_j\) carries a set-valued target support \(S_j\). For region supervision, \(S_j\) is the valid region derived from a box or mask; for point supervision, \(S_j=\{g_j\}\) is a singleton support at the annotated point. For a predicted point \(p_i\), we define its distance to \(t_j\) as
\begin{equation}
d(p_i,t_j)=d(p_i,S_j)=
\begin{cases}
0, & p_i\in S_j,\\
\inf_{u\in S_j}\|p_i-u\|_2, & p_i\notin S_j.
\end{cases}
\end{equation}
This definition follows the verifier semantics while making localization errors graded rather than all-or-nothing. For a region target, any point inside \(S_j\) is a valid hit and has zero distance, whereas an outside point is scored by its distance to the nearest point in the support. For a singleton point support, the distance reduces to the Euclidean distance to the annotated point.

Rather than using this distance as a hard accept-or-reject signal, PointRL converts it into a soft localization score:
\begin{equation}
s(p_i,t_j)
=
\exp\left(
-\frac{d(p_i,t_j)^2}{2\sigma^2}
\right).
\end{equation}
Here, \(\sigma\) is the localization bandwidth. The score is maximized on the target support and decreases smoothly as the distance grows. This gives near-miss predictions a useful learning signal while keeping the verifier target as the reward reference. The hard support-membership criterion remains \(p_i\in S_j\); the soft score determines the amount of localization credit assigned to a matched pair.

For region targets, soft distance alone can over-credit points just outside a valid box or mask, because their distance to the nearest point in \(S_j\) can be small even when they are outside the support. To preserve the distinction between a valid hit and a near miss, PointRL caps only outside-region predictions:
\begin{equation}
\bar{s}(p_i,t_j)=
\min\left(s(p_i,t_j),\tau(S_j)\right).
\end{equation}
Inside-region predictions and singleton point supports use the uncapped score \(\bar{s}(p_i,t_j)=s(p_i,t_j)\). Here, \(\tau(S_j)\in[0,1]\) denotes the cap associated with the region support type. The cap preserves dense feedback for near misses, but reserves high localization scores for points inside the target support. Mask or annotated-region supports use the highest cap, box-derived supports use an intermediate cap, and weakly localized or heuristic supports use the lowest cap. The experimental caps are fixed by support reliability as described in Section~\ref{sec:experimental-setup}.

After this step, every prediction-target pair has a comparable capped score. These scores form the matrix \(M\in\mathbb{R}^{n\times m}\):
\begin{equation}
M_{ij}
=
\bar{s}(p_i,t_j).
\end{equation}
Because neither the predicted point collection nor the target set has a canonical order, the verifier must decide which prediction should be compared with which target. The counts \(n\) and \(m\) may also differ when the response contains extra or missing points. PointRL therefore selects the best one-to-one matching with Hungarian matching~\citep{Kuhn1955TheHM,carion2020end}:
\begin{equation}
\pi^{\star}
=
\arg\max_{\pi}
\sum_{(i,j)\in\pi} M_{ij},
\end{equation}
where \(\pi\) ranges over one-to-one matchings with cardinality \(|\pi|=\min(n,m)\), so each predicted point and each target can be matched at most once.
Low-quality pairs may still be matched, but they contribute little because their \(M_{ij}\) values are low. The resulting assignment \(\pi^{\star}\) separates the response into matched prediction-target pairs and, when \(n\ne m\), unmatched predictions or missed targets. The next subsection turns these outcomes into the final reward: matched pairs provide localization credit, while extra predictions, missed targets, duplicates, and instruction-specific failures are handled by collection-level terms and guards.

\subsection{Hierarchical Pointing Reward}
\label{sec:reward_computation}

Given the matching assignment \(\pi^{\star}\), this subsection builds the response-level reward in three stages: first converting matched/unmatched outcomes into local matching quality, then adding collection-level consistency terms, and finally applying instruction-conditioned guards for parseability and robustness. The same formulation applies to both single-target and multi-target pointing. When \(m=1\), the target set is a singleton: coverage reduces to the matched target credit, while count and duplicate terms penalize extra or repeated predictions. When \(m>1\), the same terms additionally enforce coverage and cardinality consistency across the required target entries. For parseable responses, PointRL first combines matched-pair localization quality with collection-level consistency to obtain a base reward, and then applies instruction guards. The format indicator \(r_{\mathrm{fmt}}\) is determined during parsing and gates the final reward at the end; the terms below are evaluated only for parseable responses with \(n\ge 1\), and all constructed pointing samples have \(m\ge 1\).

The local reward scores the matched pairs under two normalizations. The prediction-side term measures the average matched credit per predicted point, while the target-side term measures the average matched credit per required target:
\begin{equation}
S_{\mathrm{pred}}
=
\frac{1}{n}
\sum_{(i,j)\in\pi^{\star}} M_{ij},
\qquad
S_{\mathrm{cov}}
=
\frac{1}{m}
\sum_{(i,j)\in\pi^{\star}} M_{ij}.
\end{equation}
Unmatched predictions remain in the denominator of \(S_{\mathrm{pred}}\), and unmatched target entries remain in the denominator of \(S_{\mathrm{cov}}\), so both contribute zero credit. When \(m=1\) and \(n=1\), these two terms reduce to the matched localization score for the single target. When the model returns extra points or misses required targets, the two normalizations expose these errors from the prediction and target sides, respectively. The local reward combines these two views of matched-pair quality:
\begin{equation}
R_{\mathrm{local}}
=
\lambda S_{\mathrm{pred}}+(1-\lambda)S_{\mathrm{cov}}.
\end{equation}
Here, \(\lambda\in[0,1]\) controls the balance between prediction-side and target-side credit. Thus, extra predictions reduce the prediction-side score, and missed target entries reduce the target-side coverage score.

Local matching quality alone does not ensure that the returned point collection has the required cardinality or structure. In the singleton case, a response may contain a well-localized point but also include extra points. In the multi-target case, it may localize some targets while missing others or collapse several predictions into the same local region. We therefore define global terms for count consistency and near-duplicate suppression. Count consistency compares the predicted count \(n\) with the required target count \(m\):
\begin{equation}
S_{\mathrm{cnt}}
=
\exp\left(
-\gamma_o\frac{(n-m)_+}{\max(m,1)}
-\gamma_u\frac{(m-n)_+}{\max(m,1)}
\right),
\end{equation}
where $(x)_+=\max(x,0)$. 
The coefficients $\gamma_o$ and $\gamma_u$ control the penalties for over-counting and under-counting, respectively, and we use $\gamma_u>\gamma_o$ to penalize missed targets more strongly.

To discourage local duplicate collapse, we define a duplicate suppression term:
\begin{equation}
S_{\mathrm{dup}} =
\begin{cases}
1 - \dfrac{1}{\binom{n}{2}}
\displaystyle\sum_{1 \le i < k \le n}
\mathbb{I}\!\left[\lVert p_i - p_k \rVert_2 < \delta_{\mathrm{dup}}\right],
& n \ge 2, \\[8pt]
1, & n < 2 .
\end{cases}
\label{eq:duplicate_suppression}
\end{equation}
Here, \(\delta_{\mathrm{dup}}\) is the duplicate-distance threshold in pixels, fixed at 5 pixels in our experiments. The term averages over all unordered pairs of predicted points, so the penalty is normalized by the number of possible point pairs. Duplicate coordinates are retained by the parser as separate predictions; the reward handles near-identical collapse through duplicate suppression and, when applicable, through the prediction-side, count, and coverage terms.

The global reward is defined as the product of target coverage, count consistency, and duplicate suppression:
\begin{equation}
R_{\mathrm{global}}
=
S_{\mathrm{cov}}S_{\mathrm{cnt}}S_{\mathrm{dup}}.
\end{equation}
Since $S_{\mathrm{cov}}$, $S_{\mathrm{cnt}}$, and $S_{\mathrm{dup}}$ are normalized to $[0,1]$, $R_{\mathrm{global}}$ is also bounded in $[0,1]$. 
In \(R_{\mathrm{local}}\), \(S_{\mathrm{cov}}\) contributes additively as target-side localization quality. In \(R_{\mathrm{global}}\), it acts as a soft coverage gate over the verifier target set \(T\), which may contain either one target entry or multiple entries. The global term penalizes predictions that fail to cover required targets, produce the wrong number of points, or collapse multiple predictions into a small region.

PointRL combines local matching quality and global consistency of the returned point collection as
\begin{equation}
R_{\mathrm{set}}
=
(1-\alpha)R_{\mathrm{local}}+\alpha R_{\mathrm{global}},
\end{equation}
where $\alpha\in[0,1]$ controls the balance between local point-target quality and global consistency of the returned point collection.
We set $\alpha=0.5$ in all experiments.

\begin{figure}[!t]
    \centering
    \includegraphics[width=\columnwidth]{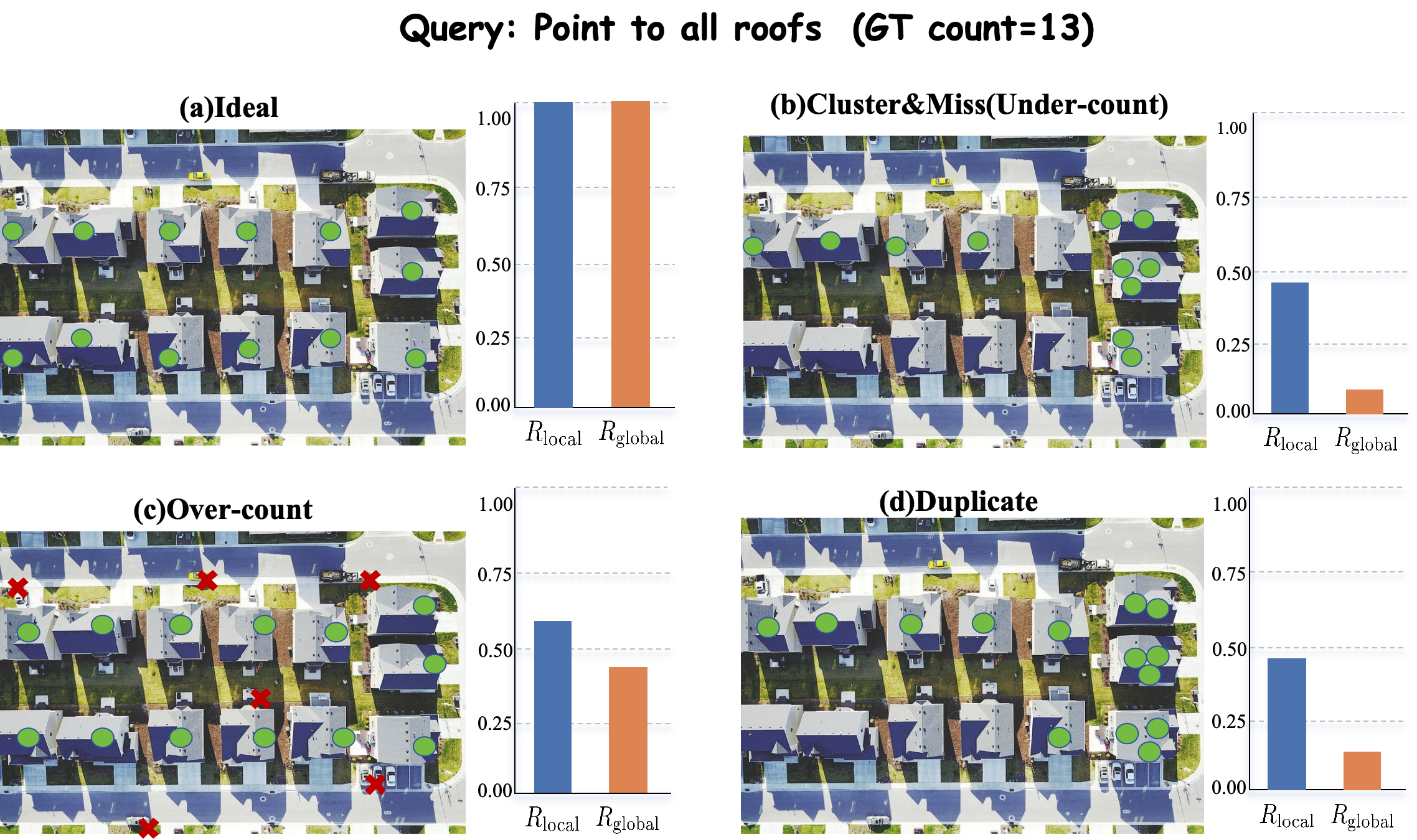}
    \caption{Local and global reward behavior for a multi-instance query. Green circles lie in target supports and red crosses lie outside. Local reward measures point-target quality, whereas global reward penalizes missed targets, incorrect counts, and collapsed predictions. The same formulation handles singleton targets.}
    \label{fig:local_global_reward}
\end{figure}

Figure~\ref{fig:local_global_reward} illustrates the multi-target failure modes that motivate the global terms; the same reward also covers the singleton case because a single target is represented as \(m=1\).

After the base reward is computed, the verifier applies the remaining instruction-specific constraints from \(C\) as guards rather than as additional matching terms. Cardinality and uniqueness constraints encoded in \(C\) are already reflected by \(m\), \(S_{\mathrm{cnt}}\), and \(S_{\mathrm{dup}}\). The indicator \(\mathbb{I}_{\mathrm{cond}}\in\{0,1\}\) is a collection-level check computed from \(P\), the matching \(\pi^\star\), and \(C\): relation-conditioned samples must satisfy the stored spatial predicate on the matched verifier entries, functional-region samples must hit one admissible functional support, and samples without such constraints set \(\mathbb{I}_{\mathrm{cond}}=1\).

Some instruction-conditioned samples contain auxiliary cue information, such as a current point, a reference object, or an anchor used to define the target. Such cues are not themselves the desired output, and they create a shortcut: a model may stay near the cue instead of selecting a point on the intended target support. We therefore treat shortcut control as a verifier-side guard.

Let \(a\) denote the stored auxiliary anchor, when present, and let \(\mathcal{S}_T=\cup_j S_j\) denote the union of target supports, where point targets are treated as singleton supports. We compute \(d(p,\mathcal{S}_T)=0\) if \(p\in\mathcal{S}_T\) and \(d(p,\mathcal{S}_T)=\inf_{u\in\mathcal{S}_T}\|p-u\|_2\) otherwise. We mark a response as cue-copying if at least one predicted point lies within \(\delta_{\mathrm{anc}}\) pixels of \(a\), while no predicted point lies inside any target support in \(\mathcal{S}_T\). This guard is activated only when the verifier evidence contains such an auxiliary cue, and is separate from \(\mathbb{I}_{\mathrm{cond}}\). All reward distances and thresholds, including \(\sigma\), \(\delta_{\mathrm{dup}}\), and \(\delta_{\mathrm{anc}}\), are computed after conversion to the verifier pixel coordinate system.

When the conditioning cue has a point anchor, we optionally retain a weak progress-shaping term:
\begin{equation}
R_{\mathrm{dir}}
=
\mathrm{clip}
\left(
\frac{d(a,\mathcal{S}_T)-\min_i d(p_i,\mathcal{S}_T)}{d(a,\mathcal{S}_T)+10^{-6}},
0,1
\right),
\end{equation}
where the denominator constant avoids division by zero. This term measures only the fractional reduction in distance to the target support relative to the cue anchor. When no point anchor is available, we set \(R_{\mathrm{dir}}=0\) and \(\lambda_{\mathrm{dir}}=0\). In the full reward, we use \(\lambda_{\mathrm{dir}}=0.10\) when this component is enabled and set \(\lambda_{\mathrm{dir}}=0\) in ablations, so the term remains subordinate to the main localization and collection-consistency rewards.

Let \(\mathbb{I}_{\mathrm{copy}}\in\{0,1\}\) indicate whether the cue-copying condition holds. The base reward after the optional shortcut cap is
\begin{equation}
R_{\mathrm{base}}
=
\begin{cases}
\min(R_{\mathrm{set}},R_{\mathrm{cap}}), & \mathbb{I}_{\mathrm{copy}}=1,\\
R_{\mathrm{set}}, & \mathbb{I}_{\mathrm{copy}}=0,
\end{cases}
\end{equation}
where \(R_{\mathrm{cap}}\in[0,1]\) is the fixed shortcut cap, set to 0.25 in our experiments. Let \(\mathbb{I}_{\mathrm{ref}}\in\{0,1\}\) indicate whether an explicit conditioning cue is present. The guarded reward applies instruction validity first, and the final format gate removes all credit from unparsable responses:
\begin{equation}
\begin{aligned}
R_{\mathrm{guard}}
&=
\mathbb{I}_{\mathrm{cond}}
\left[
(1-\mathbb{I}_{\mathrm{ref}})R_{\mathrm{set}}
+
\mathbb{I}_{\mathrm{ref}}
\min\left(1,R_{\mathrm{base}}+\lambda_{\mathrm{dir}}R_{\mathrm{dir}}\right)
\right],\\
R_{\mathrm{point}}
&=
r_{\mathrm{fmt}}R_{\mathrm{guard}},
\end{aligned}
\end{equation}
This equation defines the core verifier reward. The fixed training safeguards described in Section~\ref{sec:experimental-setup}, such as counting floors, duplicate floors, and the serialization regularizer, are used only when converting this verifier reward into the scalar reward passed to GRPO.

\subsection{Training Details}
\label{sec:training-details}

PointRL is trained on 1,647 PointRL-Data training samples with a separate 200-sample diagnostic validation split. We train with LoRA~\citep{hu2022lora}, ZeRO-2~\citep{rajbhandari2020zero}, and AdamW~\citep{loshchilov2017decoupled}, and use the hyperparameter and reward settings described in Section~\ref{sec:experimental-setup}. Model responses are serialized with a category-specific JSON contract that is mapped back to the standard point collection and rescaled to image pixels. We use a 4:1 non-counting-to-counting task-batched sampling schedule before gradient accumulation, and serialization and safety safeguards are fixed across all reported configurations.
The final pointing reward is computed after discrete text generation and cannot be backpropagated through the verifier. PointRL therefore uses it as response-level feedback for GRPO-style policy optimization~\citep{shao2024deepseekmath}.
For each training input $z=(I,q)$, the old policy samples a group of $G$ responses $\{y_i\}_{i=1}^{G}$.
Each response is parsed into a point collection \(P_i\) and evaluated by the hidden verifier evidence. Let \(\phi_{\mathrm{safe}}\) denote the fixed count and duplicate safeguards, and let \(L_{\mathrm{ans}}\) be the number of answer-field tokens used by the serialized response. The scalar reward used for GRPO is
\begin{equation}
\begin{aligned}
R_{\mathrm{train}}(P_i,T,C)
&=
\max\Bigl(
0,\,
\phi_{\mathrm{safe}}\!\left(
R_{\mathrm{point}}(P_i,T,C),\right.\\
&\qquad\left.
P_i,T,C
\right)
-
\eta_{\mathrm{ser}}L_{\mathrm{ans}}
\Bigr),
\end{aligned}
\end{equation}
where \(\eta_{\mathrm{ser}}=2.0\times10^{-4}\). The safeguard map \(\phi_{\mathrm{safe}}\) applies the fixed counting, duplicate, and hard over-count floors: responses with severe count mismatch, repeated points on the same target, or excessive unmatched predictions are capped at the corresponding fixed floor. These settings are held constant across the reported reward configurations. The policy-ratio clipping used by GRPO is applied later in the optimization objective, not as reward post-processing. We then write
\begin{equation}
r_i = R_{\mathrm{train}}(P_i,T,C).
\end{equation}
The rewards are normalized within the response group to obtain the advantage of each sample:
\begin{equation}
\hat{A}_i
=
\frac{r_i-\mu_r}{\sigma_r+\delta},
\qquad
\mu_r=\frac{1}{G}\sum_{i=1}^{G}r_i,
\end{equation}
where $\sigma_r$ is the standard deviation of group rewards and $\delta$ is a small constant for numerical stability.

The model is then updated with the clipped GRPO objective, which follows the clipped policy-ratio design of PPO~\citep{schulman2017proximal}:
\begin{equation}
\begin{split}
J(\theta)=\mathbb{E}\Bigg[
&\frac{1}{G}\sum_{i=1}^{G}\frac{1}{|y_i|}
\sum_{t=1}^{|y_i|}
\Big(
\min\Big(
\rho_{i,t}\hat{A}_i,\\
&\mathrm{clip}(\rho_{i,t},1-\epsilon,1+\epsilon)\hat{A}_i
\Big)
-\beta D_{\mathrm{KL}}^{i,t}
\Big)
\Bigg],
\end{split}
\end{equation}
where
\begin{equation}
\rho_{i,t}
=
\frac{
\pi_{\theta}(y_{i,t}\mid z,y_{i,<t})
}{
\pi_{\theta_{\mathrm{old}}}(y_{i,t}\mid z,y_{i,<t})
}.
\end{equation}
Here, $\pi_{\theta}$ denotes the current VLM policy, $\pi_{\theta_{\mathrm{old}}}$ denotes the sampling policy before the update, and $\pi_{\mathrm{ref}}$ denotes the reference policy. The term \(D_{\mathrm{KL}}^{i,t}\) denotes the token-level divergence between \(\pi_{\theta}(\cdot\mid z,y_{i,<t})\) and \(\pi_{\mathrm{ref}}(\cdot\mid z,y_{i,<t})\).
In our experiments, we set \(\beta=0\), so the KL term is disabled; it is retained in the objective to make the general GRPO form explicit.
Since $R_{\mathrm{point}}$ is computed from the complete parsed response, the same response-level advantage $\hat{A}_i$ is applied to all tokens in $y_i$. 
This optimization increases the likelihood of responses that are parseable, well localized, collection-consistent, and compliant with instruction-conditioned constraints.

\section{Experiments}
\label{sec:experiments}

We evaluate PointRL on PointArena Point-Bench and external spatial grounding benchmarks. The experiments are organized around three questions: whether verifier-based point-level training improves the base VLM under a same-backbone comparison, how the tested reward configurations behave, and whether the resulting model shows same-backbone gains on external benchmarks without benchmark-specific fine-tuning.
\subsection{Experimental Setup}
\label{sec:experimental-setup}

\textbf{Datasets and metrics.}
We evaluate on PointArena Point-Bench~\citep{cheng2025pointarena}, which covers affordance, counting, reasoning, spatial, and steerable pointing tasks. PointRL is trained on 1,647 constructed samples from the rule families and verifier evidence summarized in Table~\ref{tab:data-composition}, uses a separate 200-sample validation set as a held-out diagnostic split, and is evaluated on the 982-sample Point-Bench test set. Before training, we screened the training and validation splits against the Point-Bench test set and the external RoboSpatial~\citep{song2025robospatial}, BLINK~\citep{fu2024blink}, and Ref-Adv~\citep{dong2026ref} benchmarks by exact matching over image identifiers or hashes, source annotation identifiers, and normalized query strings when available; no matched evaluation sample is retained for reward training. Evaluation images, labels, and benchmark-specific annotations are not used for reward construction, prompt tuning, parser tuning, reward-weight selection, or checkpoint selection. We report the official hard hit-rate metric: a single-target prediction must fall inside the target mask, and a counting or multi-target point collection must have the correct count with all points inside target masks. The reported metric is binary, while PointRL training uses the denser hidden-verifier reward defined in Section~\ref{sec:reward_computation}.

\textbf{Baselines and comparison protocol.}
We report two types of comparisons. 
First, we include representative VLMs and pointing models with comparable Point-Bench results as external references. These reported numbers provide context rather than controlled evidence, since the models may differ in backbone, training data, and prompting.
Second, we conduct controlled comparisons on Qwen3.5~\citep{qwen3.5} backbones. 
For both Qwen3.5-2B and Qwen3.5-4B, we compare the base model with its PointRL-trained counterpart under the same evaluation protocol. These comparisons test the complete PointRL training recipe on the evaluated backbones, rather than isolating the verifier reward from fixed training choices such as data construction, serialization, batching, and LoRA optimization. For each benchmark, the compared Qwen3.5 rows use the same prompt, decoding setting, parser, and scorer.

\textbf{Implementation details.}
Section~\ref{sec:training-details} gives the concrete optimization and serialization settings. As noted there, those settings are fixed across the reported reward comparisons.
All PointRL variants use Qwen3.5 backbones, with Qwen3.5-4B as the main base model and Qwen3.5-2B as the smaller-scale run. Model responses follow a category-specific JSON contract that is normalized back to the standard point collection \(P\); coordinates are serialized on \([0,1000]\) and rescaled to image pixels for verification. GRPO training uses \(G=4\), clipping range \(\epsilon=0.2\), KL coefficient \(\beta=0\), 1,000 update steps, and a maximum completion length of 512 tokens. Optimization uses LoRA with \(r=32\), \(\alpha=64\), dropout \(0.05\), learning rate \(5.0\times10^{-6}\), bf16 precision, ZeRO-2, AdamW, and maximum gradient norm \(1.0\). Task-batched sampling follows a 4:1 non-counting-to-counting schedule before gradient accumulation.

The reward settings are also fixed across the reported reward comparisons. We set \(\sigma_{\mathrm{cnt}}=50\) pixels for counting rewards and \(\sigma_{\mathrm{mask}}=40\) pixels for non-counting and steerable rewards. The local and set balance weights are \(\lambda=0.5\) and \(\alpha=0.5\), with under-count and over-count penalties \(\gamma_u=3.0\) and \(\gamma_o=1.2\). Duplicate and hard over-count floors are set to \(0.3\), region-validity caps use \(\tau(S_j)\in\{0.70,0.55,0.35\}\) by support reliability, and the duplicate threshold is \(\delta_{\mathrm{dup}}=5\) pixels. For cue-bound instructions, the cue-copying radius is \(\delta_{\mathrm{anc}}=\max(12,\;0.15\,d(a,\mathcal{S}_T))\), with cap \(R_{\mathrm{cap}}=0.25\) and progress weight \(\lambda_{\mathrm{dir}}=0.10\) when enabled. The serialization regularizer is \(2.0\times10^{-4}\) per answer-field token.

Unless otherwise stated, each trained configuration is evaluated from a single checkpoint. We therefore interpret small differences cautiously and focus on same-backbone comparisons under the same evaluation protocol. Because the reported metrics are binary hit rates and per-example agreement statistics are not reported, we do not claim statistical significance for category-level, ablation-level, or external-benchmark differences.
	\subsection{Main Results}
	\label{sec:main-results}
\begin{figure*}[!t]
    \centering
    \includegraphics[width=\textwidth]{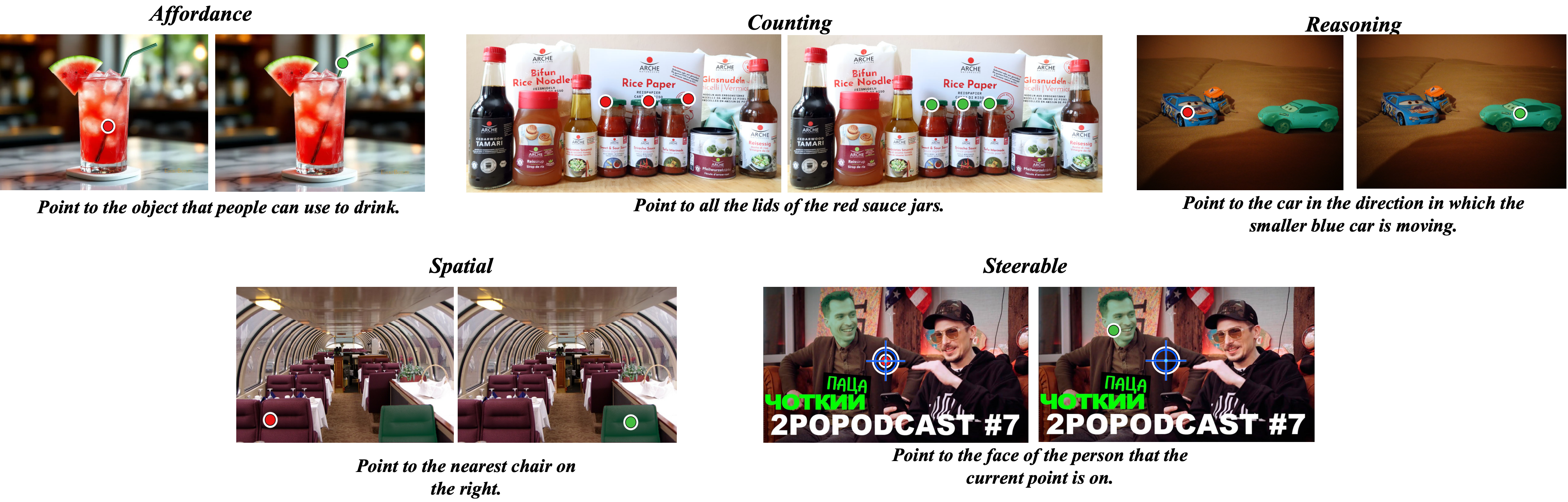}
\caption{Qualitative comparison on PointArena Point-Bench. Each category panel shows Qwen3.5-4B on the left and Qwen3.5-4B + PointRL on the right. Red and green circular markers denote the corresponding baseline and PointRL predictions; blue rings mark reference or current points in steerable examples, and translucent green regions indicate target supports when shown.}
\label{fig:main-qualitative}
\end{figure*}

\begin{table*}[!t]
\centering
\caption{Main results on PointArena Point-Bench.}
\label{tab:main_results}
\small
\newcommand{\deltarow}[8]{%
\rowcolor{DeltaRow}
#1 & #2 & #3 & #4 & #5 & #6 & #7 & #8
}
\setlength{\tabcolsep}{1.0pt}
\renewcommand{\arraystretch}{0.94}
\resizebox{\textwidth}{!}{%
\begin{tabular}{@{}>{\RaggedRight\arraybackslash}p{0.25\textwidth}
    >{\centering\arraybackslash}p{0.107142857\textwidth}
    >{\centering\arraybackslash}p{0.107142857\textwidth}
    >{\centering\arraybackslash}p{0.107142857\textwidth}
    >{\centering\arraybackslash}p{0.107142857\textwidth}
    >{\centering\arraybackslash}p{0.107142857\textwidth}
    >{\centering\arraybackslash}p{0.107142857\textwidth}
    >{\centering\arraybackslash}p{0.107142857\textwidth}}
\toprule
Method & Params & Affordance & Counting & Reasoning & Spatial & Steerable & Overall Acc \\
\midrule
GPT-4o & N/D & 42.42 & 31.12 & 23.83 & 25.64 & 24.50 & 29.50 \\
Qwen2.5-VL-7B-Instruct & 7B & 76.77 & 52.55 & 53.89 & 61.54 & 36.50 & 56.25 \\
Qwen2.5-VL-72B-Instruct & 72B & 76.77 & 57.14 & 54.40 & 60.00 & 46.50 & 58.96 \\
Molmo-72B & 72B & 87.88 & 54.59 & 69.43 & 70.26 & 37.00 & 63.83 \\
Gemini-Robotics-ER-1.5 & N/D & 69.70 & 68.47 & 60.10 & 69.74 & 67.50 & 67.10 \\
\midrule
Qwen3.5-2B & 2B & 34.34 & 10.20 & 32.12 & 33.85 & 16.00 & 25.25 \\
Qwen3.5-2B + PointRL & 2B & 64.65 & 55.10 & 58.55 & 63.08 & 43.00 & 56.82 \\
\deltarow{\quad $\Delta$ over 2B base}{--}{\deltaval{+30.31}}{\deltaval{+44.90}}{\deltaval{+26.43}}{\deltaval{+29.23}}{\deltaval{+27.00}}{\deltaval{+31.57}} \\
\specialrule{0.08em}{0pt}{0pt}
Qwen3.5-4B & 4B & 65.66 & 47.45 & 58.55 & 69.23 & 40.00 & 56.11 \\
Qwen3.5-4B + PointRL & 4B & 75.25 & 60.71 & 70.47 & 71.28 & 50.50 & 65.58 \\
\deltarow{\quad $\Delta$ over 4B base}{--}{\deltaval{+9.59}}{\deltaval{+13.26}}{\deltaval{+11.92}}{\deltaval{+2.05}}{\deltaval{+10.50}}{\deltaval{+9.47}} \\
\bottomrule
\end{tabular}}
\end{table*}

Table~\ref{tab:main_results} reports the main results on PointArena Point-Bench. All values are percentages, and Overall Acc is micro-average accuracy. The Params column reports disclosed model sizes, while N/D marks models whose parameter counts are not disclosed. In the controlled same-backbone comparison, PointRL improves Qwen3.5-4B from 56.11\% to 65.58\% overall, an observed gain of 9.47 percentage points under the single-checkpoint protocol. With the same reward settings, Qwen3.5-2B improves from 25.25\% to 56.82\%. These results indicate that PointRL training yields gains over both evaluated base backbones in this setup.

The gains are especially pronounced on tasks that require either collection structure or instruction-conditioned target selection. For Qwen3.5-4B, the largest category gains are on counting, reasoning, and steerable tasks, with improvements of 13.26, 11.92, and 10.50 points, respectively. Affordance also improves by 9.59 points, whereas spatial improves by 2.05 points from a stronger baseline score of 69.23\%. This pattern suggests that the verifier reward is most useful when the model must decide not only where to point, but also how many targets to return and which constraint-bound target set is valid. The Qwen3.5-2B run shows a larger overall gain of 31.57 points, indicating that the same reward design can substantially improve a weaker base model, although the trained 4B model remains stronger overall.

The qualitative examples in Fig.~\ref{fig:main-qualitative} complement the numerical results by showing how the trained model changes the predicted point locations. In the shown cases, PointRL moves predictions toward instruction-relevant regions and improves target coverage in multi-target examples, which is consistent with the stronger gains on counting, reasoning, and steerable categories.

\subsection{Reward Ablation}
\label{sec:reward-ablation}
\begin{figure*}[!t]
\centering
\includegraphics[width=\textwidth]{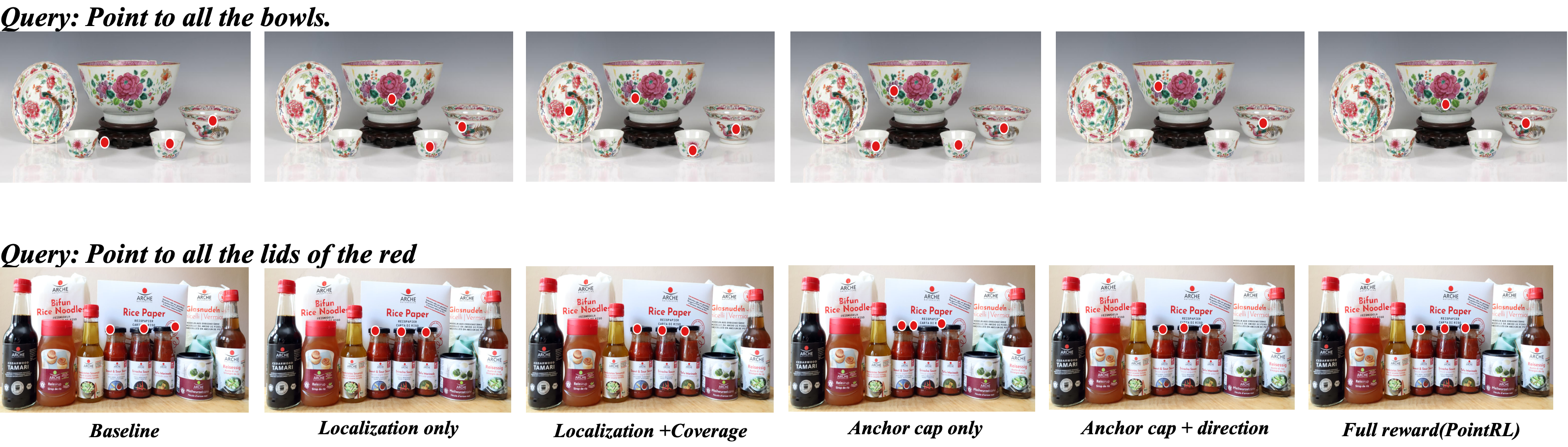}
    \caption{Comparison of reward configurations on multi-target pointing examples. Each row is one query; columns show the pretrained baseline, localization-only reward, localization plus target coverage, cue-copying shortcut cap, cue-copying shortcut cap with progress shaping, and the full PointRL reward. Red points denote predicted point collections.}
\label{fig:reward-ablation-qualitative}
\end{figure*}

Table~\ref{tab:reward_ablation} compares selected reward configurations under the same base model and evaluation protocol. The ablation is configuration-level: it reports how the tested reward bundles behave under a shared setup rather than isolating one reward term at a time. In this table, the cue-copying shortcut cap limits predictions that stay near an auxiliary cue while missing the target, and the progress variant additionally uses the weak progress term defined in Section~\ref{sec:reward_computation}.
For compactness, Loc. denotes localization reward, Cov. denotes target coverage, Cue cap denotes the cue-copying shortcut cap, and Prog. denotes progress shaping.

\begin{table*}[!t]
\centering
\caption{Reward ablation on PointArena Point-Bench.}
\label{tab:reward_ablation}
\footnotesize
\setlength{\tabcolsep}{4pt}
\renewcommand{\arraystretch}{1.08}
\begin{tabular*}{\textwidth}{@{\extracolsep{\fill}}lcccccc@{}}
\toprule
Method & Affordance & Counting & Reasoning & Spatial & Steerable & Overall Acc \\
\midrule
Qwen3.5-4B & 65.66 & 47.45 & 58.55 & 69.23 & 40.00 & 56.11 \\
+ Loc. only & 69.19 & 59.18 & 60.62 & 69.23 & 43.00 & 60.18 \\
+ Loc. + Cov. & 67.17 & 57.14 & 68.91 & \textbf{71.28} & 48.00 & 62.42 \\
+ Cue cap & 67.68 & 55.10 & 66.32 & 68.72 & 49.50 & 61.41 \\
+ Cue cap + Prog. & 69.70 & 56.63 & 66.84 & 66.15 & 41.00 & 59.98 \\
+ Full reward & \textbf{75.25} & \textbf{60.71} & \textbf{70.47} & \textbf{71.28} & \textbf{50.50} & \textbf{65.58} \\
\bottomrule
\end{tabular*}
\end{table*}

In this single-run configuration comparison, the full reward obtains the highest observed overall accuracy, 65.58\%, which is 9.47 points above the base model and at least 3.16 points above the tested partial reward variants. The localization-only reward improves overall accuracy to 60.18\%, showing that distance-based feedback already provides a useful signal beyond binary hit checking. Adding target coverage raises the overall score to 62.42\%, a further 2.24-point gain over localization only, and mainly improves reasoning, spatial, and steerable tasks. The cue-cap configurations also improve over the base model, but their behavior is less stable: the progress variant increases affordance slightly while reducing steerable performance and overall accuracy relative to cue cap alone. These trends support the use of the full reward bundle, where local localization, target coverage, count consistency, duplicate suppression, and guard terms act together rather than as isolated heuristics.

Overall, the table should be read as configuration-level evidence rather than as a causal study of individual reward terms. Among the tested bundles, the full reward is the strongest observed configuration. Because count consistency and duplicate suppression are not isolated in separate rows, their marginal contributions would require additional controlled variants and repeated runs. Fig.~\ref{fig:reward-ablation-qualitative} illustrates this pattern on multi-target examples: partial rewards often move predictions toward relevant regions but still leave missing, collapsed, or duplicated points, whereas the full reward better covers the target set in the shown examples.

\subsection{External Benchmarks and Distractor-Stratified Analysis}
\label{sec:generalization-robustness}

\begin{figure*}[!t]
\centering
\includegraphics[width=0.90\textwidth]{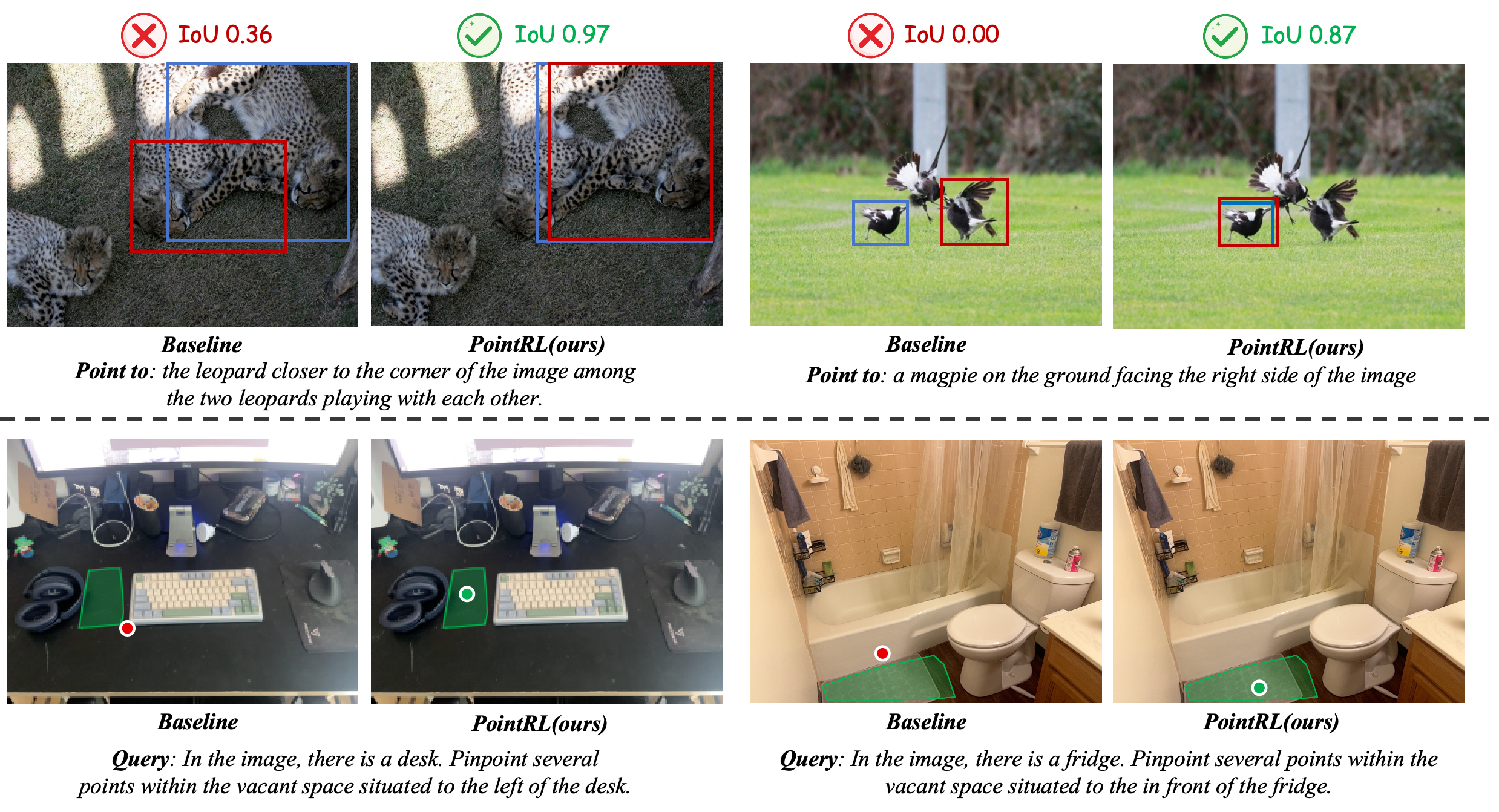}
\caption{External-benchmark examples on Ref-Adv and RoboSpatial-Home. In each pair, the left panel shows Qwen3.5-4B and the right panel shows Qwen3.5-4B + PointRL. In the Ref-Adv row, red boxes are predictions, blue boxes are targets, and IoU is reported above each panel. In the RoboSpatial-Home row, green regions mark valid target regions; red and green dots denote baseline and PointRL predictions, respectively.}
\label{fig:external-qualitative-1}
\end{figure*}

\begin{figure*}[!t]
\centering
\includegraphics[width=\textwidth]{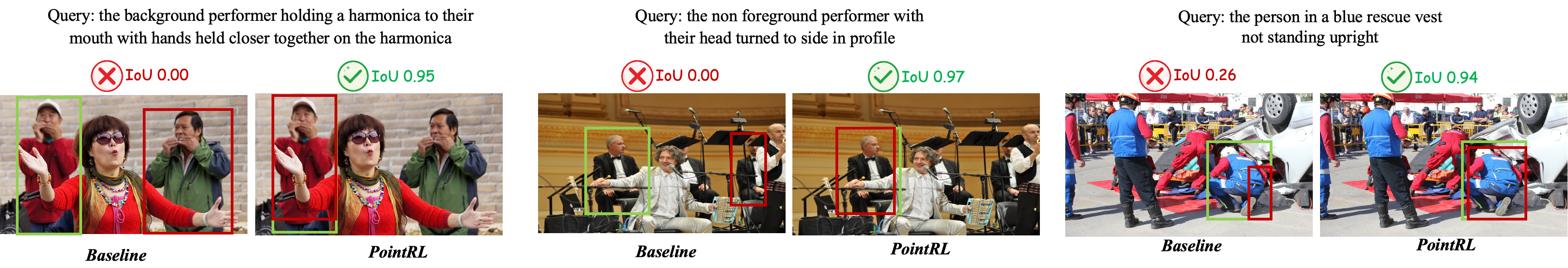}
\caption{Ref-Adv predictions under increasing distractor counts. The three example groups correspond to 2--3, 4--6, and \(\geq 7\) distractors from left to right; within each group, the left panel is Qwen3.5-4B and the right panel is Qwen3.5-4B + PointRL. Red boxes denote predictions, blue boxes denote targets, and IoU is reported above each panel.}
\label{fig:ref-adv-distractor}
\end{figure*}

We further evaluate PointRL on RoboSpatial~\citep{song2025robospatial}, BLINK~\citep{fu2024blink}, and Ref-Adv~\citep{dong2026ref} as external benchmarks. RoboSpatial and BLINK test spatial grounding directly, whereas Ref-Adv evaluates target selection through its required box-output protocol. PointRL training uses a separate training set from these benchmarks, and the Qwen3.5-4B + PointRL row is evaluated without benchmark-specific fine-tuning. Benchmark-trained methods are included only as in-domain references. On Ref-Adv, both Qwen3.5-4B rows use the same box prompt, invalid-output handling, decoding setting, parser, and benchmark-required output format, all fixed before evaluation; the model is therefore evaluated through the official box-output interface rather than through the point parser used during PointRL reward training. This evaluation tests transfer to target selection under the Ref-Adv protocol, while leaving the point-reward formulation used for PointRL training unchanged.
\begin{table}[!htbp]
\centering
\caption{External evaluation on RoboSpatial and BLINK. All values are percentages under the official benchmark metrics. The three RoboSpatial columns report the Context, Configuration, and Compatibility splits. Methods marked with $^{+}$ use benchmark-specific training data.}
\label{tab:robospatial_blink}
\footnotesize
\setlength{\tabcolsep}{4pt}
\renewcommand{\arraystretch}{1.05}
\begin{tabularx}{\linewidth}{@{}>{\RaggedRight\arraybackslash}Xcccc@{}}
\toprule
Method & Context & Config. & Compat. & BLINK \\
\midrule
Molmo & 0.00 & 48.70 & 24.80 & 67.10 \\
GPT-4o & 6.60 & 74.00 & 55.20 & 76.20 \\
RoboPoint & 2.50 & 59.40 & 80.10 & 63.60 \\
RoboPoint + RoboSpatial$^{+}$ & 21.30 & 70.00 & 88.60 & 70.60 \\
LLaVA-NeXT + RoboSpatial$^{+}$ & 19.70 & 76.40 & 80.10 & 79.00 \\
\midrule
Qwen3.5-4B (baseline) & 27.05 & 79.67 & 25.71 & 66.43 \\
Qwen3.5-4B + PointRL (ours) & 34.43 & 86.90 & 39.05 & 73.43 \\
\bottomrule
\end{tabularx}
\end{table}

Table~\ref{tab:robospatial_blink} shows observed gains over the Qwen3.5-4B baseline across the evaluated external metrics under this protocol. On the RoboSpatial context, configuration, and compatibility splits, and on BLINK, PointRL brings gains of 7.38, 7.23, 13.34, and 7.00 percentage points, respectively. These results provide bounded evidence of cross-benchmark improvement in the evaluated settings. In-domain RoboSpatial-trained models remain stronger on compatibility, indicating that benchmark-specific supervision still provides additional benefits. Illustrative Ref-Adv and RoboSpatial-Home examples are shown in Figure~\ref{fig:external-qualitative-1}.

The external benchmark results also clarify what transfers and what does not. The compatibility split has the largest same-backbone improvement, but the absolute score of Qwen3.5-4B + PointRL remains below methods trained directly with RoboSpatial data. This suggests that PointRL improves general spatial grounding behavior, while benchmark-specific supervision still matters for the compatibility distribution. The BLINK gain is useful for the opposite reason: it is obtained without BLINK-specific tuning, so it supports the claim that verifier-based point feedback can transfer beyond the constructed PointRL-Data distribution. We therefore treat these results as evidence of controlled same-backbone improvement, not as a claim that PointRL replaces benchmark-specific training.

\begin{table}[!t]
\centering
\caption{Ref-Adv box-output evaluation under IoU thresholds and distractor-count splits. All values are percentages.}
\label{tab:ref_adv}
\footnotesize
\setlength{\tabcolsep}{3.5pt}
\renewcommand{\arraystretch}{1.02}
\begin{tabularx}{\linewidth}{@{}>{\RaggedRight\arraybackslash}Xcccccc@{}}
\toprule
Method & Acc@0.5 & Acc@0.75 & Acc@0.9 & 2--3 & 4--6 & \(\geq 7\) \\
\midrule
Qwen2.5-VL-72B & 54.00 & 40.10 & 18.00 & 57.00 & 52.70 & 41.10 \\
Qwen3-VL-8B & 52.30 & 38.90 & 19.90 & 55.70 & 50.20 & 38.80 \\
Qwen3-VL-32B & 53.40 & 44.70 & 27.10 & 56.30 & 50.20 & 45.70 \\
Qwen3-VL-30B-A3B & 52.10 & 43.10 & 27.40 & 54.70 & 48.90 & 45.70 \\
Qwen3-VL-4B-Thinking & 57.60 & 45.50 & 27.80 & 63.00 & 52.70 & 40.30 \\
\midrule
Qwen3.5-4B (baseline) & 52.28 & 37.83 & 21.10 & 55.16 & 49.84 & 42.64 \\
Qwen3.5-4B + PointRL & 54.29 & 41.16 & 23.03 & 56.45 & 53.33 & 44.96 \\
\bottomrule
\end{tabularx}
\end{table}

Table~\ref{tab:ref_adv} reports Ref-Adv results under IoU thresholds and distractor-count strata. Under our same-backbone evaluation protocol, PointRL has higher observed scores than Qwen3.5-4B on all six reported metrics. The gains are modest: 2.01, 3.33, and 1.93 points under the three IoU thresholds, and 1.29, 3.49, and 2.32 points under the 2--3, 4--6, and $\geq 7$ distractor settings. Since Ref-Adv is a box-output benchmark, we interpret these results as bounded evidence of target-selection transfer under the Ref-Adv protocol, not as a direct point-output result or a standalone claim about distractor robustness. Qualitative examples in Fig.~\ref{fig:external-qualitative-1} show Ref-Adv target selection and RoboSpatial-Home free-space localization cases. We further visualize Ref-Adv examples under increasing distractor counts in Fig.~\ref{fig:ref-adv-distractor}, covering the 2--3, 4--6, and $\geq 7$ distractor settings.

The Ref-Adv improvements are smaller than the Point-Bench gains, which is expected because the evaluation uses the benchmark-required box-output interface rather than the point parser used during reward training. Even so, the gains are consistent across strictness levels. The improvement at Acc@0.75 is larger than at Acc@0.5, and the 4--6 distractor stratum has the largest distractor-based gain. This pattern is compatible with improved target selection rather than a simple formatting advantage: PointRL does not change the Ref-Adv output protocol, but it can make the model choose the intended object more reliably when the scene contains competing distractors.

Across the external evaluations, the same-backbone comparison yields positive gains under both point- and box-output protocols, but the magnitude varies by benchmark and split. On RoboSpatial, the largest observed gain is on compatibility (+13.34), with gains of +7.38 and +7.23 on context and configuration, respectively. The +7.00 improvement on BLINK provides additional evidence outside the constructed PointRL-Data distribution. On Ref-Adv, the gains across the three distractor strata range from 1.29 to 3.49 points, with the largest observed improvement in the 4--6 range. Because Ref-Adv is evaluated through its required box-output interface, these results are best read as transfer to target selection under a different output protocol. The lower compatibility score relative to RoboSpatial-trained methods and the single-checkpoint setting limit the claim to same-backbone gains.

\FloatBarrier
\section{Conclusion}
\label{sec:conclusion}

This work presents PointRL, a verifiable reinforcement learning framework for learning point-level visual grounding from heterogeneous grounding annotations. PointRL converts boxes, masks, and instance labels into pointing instructions and keeps the original annotations as hidden verifier evidence for reward computation. This design preserves non-unique valid coordinates for region-based supports and allows the reward to credit different points that satisfy the annotated target condition.

PointRL further defines a structured pointing reward over unordered point collections. The reward combines parseability, soft localization, target coverage, count consistency, duplicate suppression, and constraint checks. In this way, it supports both single-target and multi-target grounding under a unified formulation. Experiments on PointArena yield an observed 9.47-point gain over the Qwen3.5-4B baseline in this protocol, and the configuration-level ablations suggest that local localization feedback and collection-level structural constraints are useful in combination in the tested settings. Additional evaluations on RoboSpatial, BLINK, and Ref-Adv show bounded same-backbone gains under the evaluated protocols. These results suggest that, in the evaluated settings, heterogeneous grounding annotations can serve as useful hidden verifier evidence for improving the pointing ability of VLMs.

\balance
\bibliographystyle{IEEEtran}
\bibliography{references}

\begin{IEEEbiography}[{\includegraphics[width=1in,height=1.25in,clip,keepaspectratio]{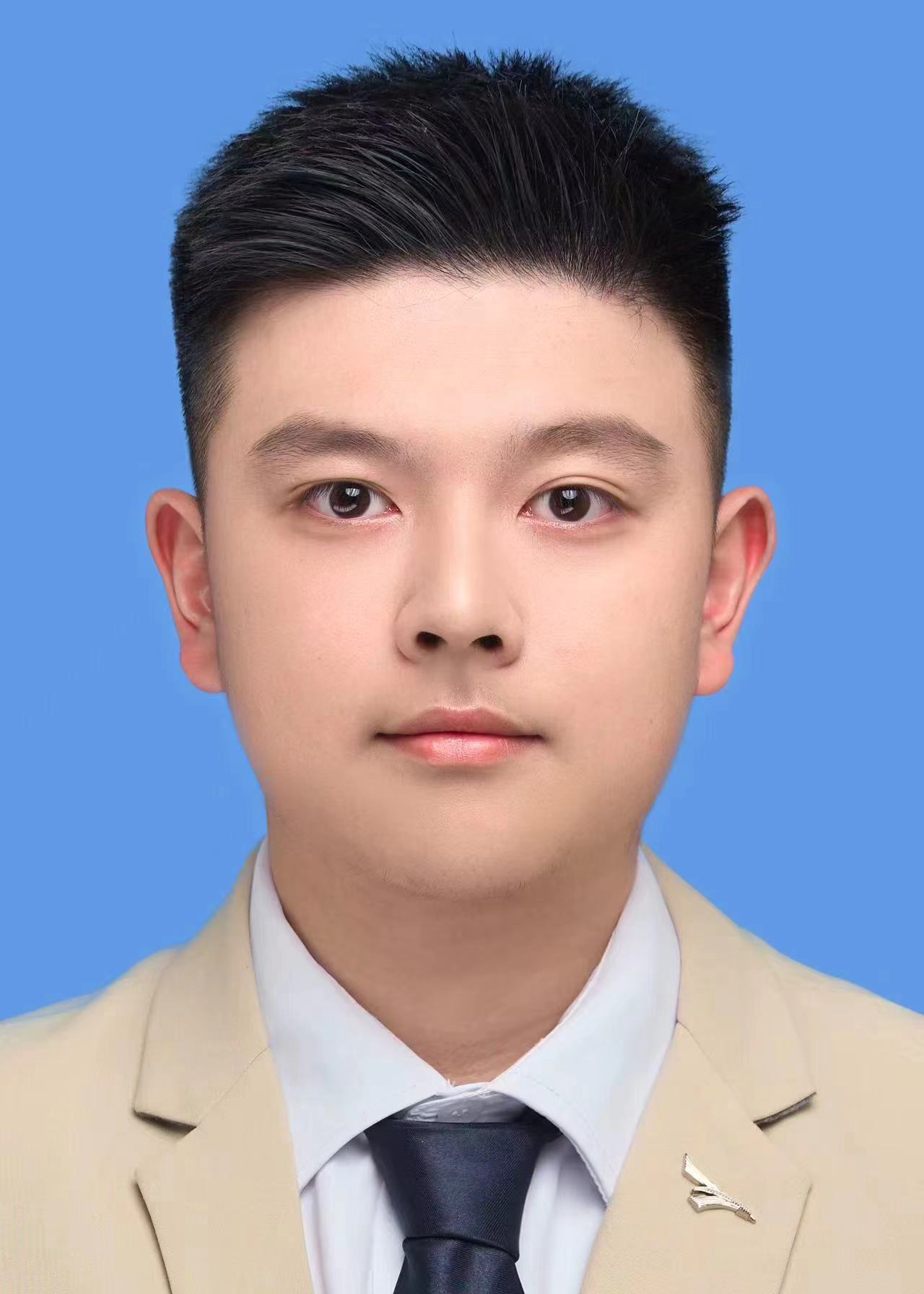}}]{Jingyang Su}
received the B.E. degree in automation from Hangzhou Dianzi University, Hangzhou, China, in 2025. He is currently pursuing the M.S. degree with the School of Intelligent Engineering and Automation, Beijing University of Posts and Telecommunications (BUPT), Beijing, China. His research interests include multimodal large language models, visual grounding, reinforcement learning, and uncertainty estimation.
\end{IEEEbiography}
\vspace{-10mm}

\begin{IEEEbiography}[{\includegraphics[width=1in,height=1.25in,clip,keepaspectratio]{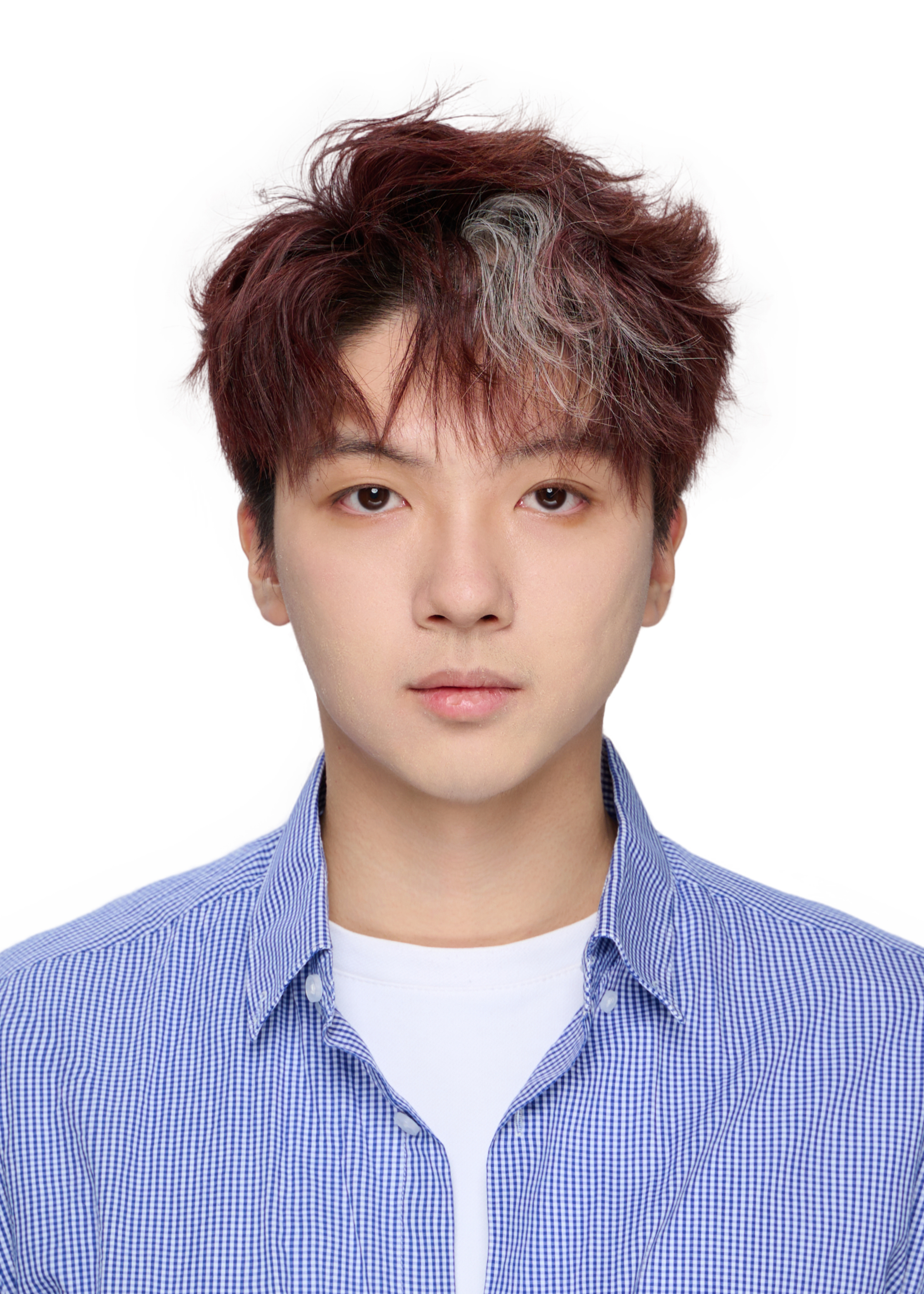}}]{Pu Cao}
received his bachelor's degree from the University of Science and Technology Beijing (USTB), Beijing, China, in 2022, and is currently a Ph.D. candidate at the School of Intelligent Engineering and Automation, Beijing University of Posts and Telecommunications (BUPT), Beijing, China, since 2022. His research interests include multimodal understanding and generation, especially multimodal large language models and diffusion models.
\end{IEEEbiography}
\vspace{-10mm}

\begin{IEEEbiography}[{\includegraphics[width=1in,height=1.25in,clip,keepaspectratio]{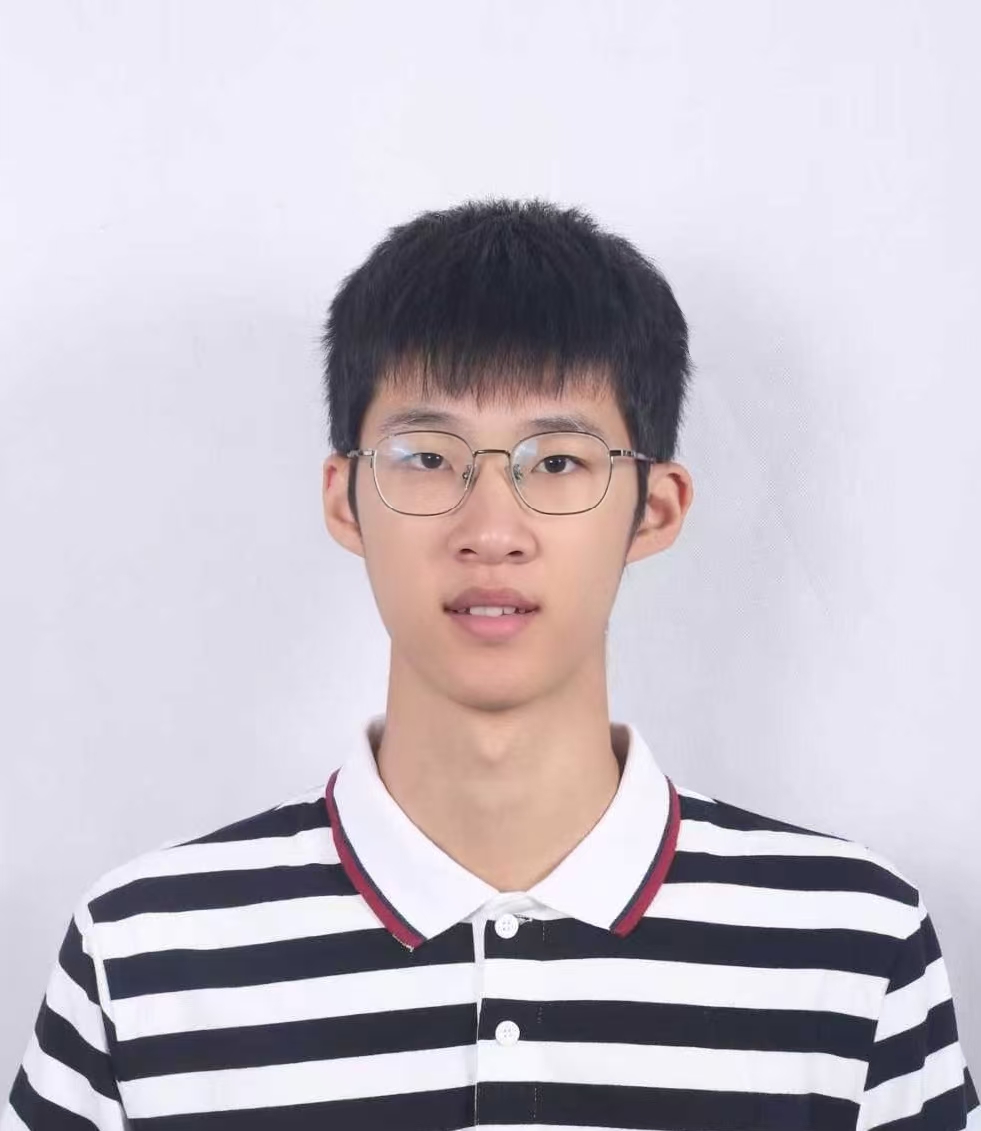}}]{Xiuze Jin}
entered Beijing University of Posts and Telecommunications (BUPT) in 2025 and is currently majoring in the Computer Science Meta-Class of the Future College. During his freshman year, he developed a strong interest in large AI models and robotics and is eager to pursue research in these fields.
\end{IEEEbiography}
\vspace{-10mm}

\begin{IEEEbiography}[{\includegraphics[width=1in,height=1.25in,clip,keepaspectratio]{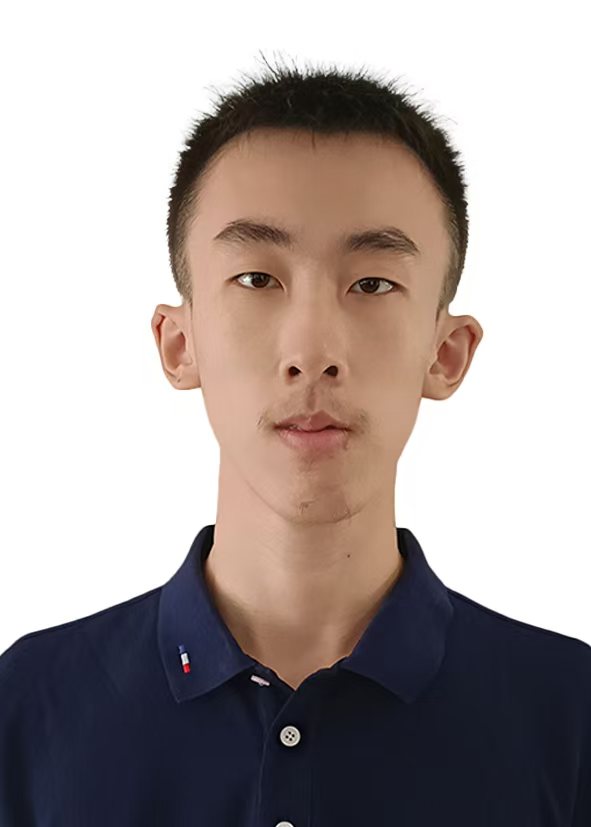}}]{Longyue Zhang}
is currently pursuing his undergraduate studies in automation at Beijing University of Posts and Telecommunications (BUPT), Beijing, China. His research interests lie at the intersection of mechanics and control systems, as well as human--large language model interaction. He is committed to rigorous and collaborative interdisciplinary research.
\end{IEEEbiography}
\vspace{-10mm}

\begin{IEEEbiography}[{\includegraphics[width=1in,height=1.25in,clip,keepaspectratio]{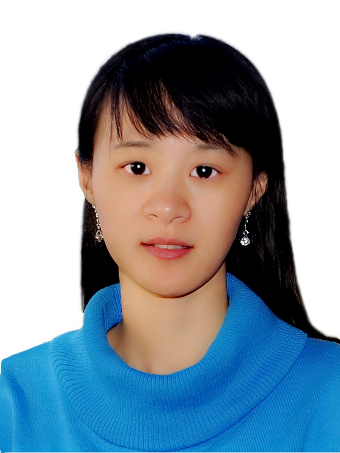}}]{Qing Song}
received the Ph.D. degree from Tianjin University, Tianjin, China, in 2006. She is currently a researcher with Beijing University of Posts and Telecommunications (BUPT), where she is engaged in computer vision technology. She founded the Pattern Recognition and Intelligent Vision Laboratory (PRIV).
\end{IEEEbiography}
\vspace{-10mm}

\begin{IEEEbiography}[{\includegraphics[width=1in,height=1.25in,clip,keepaspectratio]{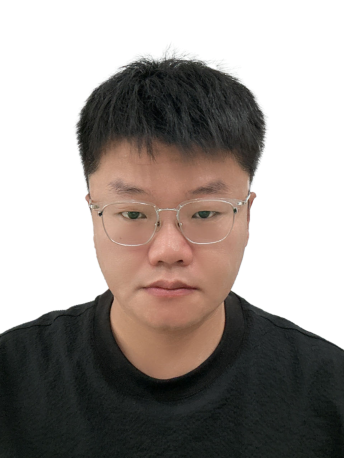}}]{Lu Yang}
is currently an associate professor with Beijing University of Posts and Telecommunications (BUPT), China. He received the Ph.D. degree from BUPT in 2021. He has conducted research with the Pattern Recognition and Intelligent Vision Laboratory (PRIV) since 2012. His research interests include human-centric AI and generative AI.
\end{IEEEbiography}

\end{document}